%% file: main.tex
\documentclass[
]{ceurart}

\usepackage{dirtytalk}
\usepackage{tabularx}
\usepackage{listings}
\usepackage{adjustbox}
\begin{document}

\copyrightyear{2022}
\copyrightclause{Copyright for this paper by its authors.
  Use permitted under Creative Commons License Attribution 4.0
  International (CC BY 4.0).}

%\conference{19th IRCDL (The Conference on Information and Research science Connecting to Digital and Library science), February 23--24, 2023, Bari, Italy}
\conference{}

\title{Grimm in Wonderland: Prompt Engineering with Midjourney to Illustrate Fairytales}

%\title{Grimm Hallucinations: Prompt Engineering with Midjourney to Illustrate Fairytales}
% \tnotetext[1]{You can use this document as the template for preparing your
%   publication. We recommend using the latest version of the ceurart style.}

\author{Martin Ruskov}[%
%email=martin.ruskov@unimi.it,
% url=https://islab.di.unimi.it/team/martin.ruskov@unimi.it,
%orcid=0000-0001-5337-0636
]
%% \cormark[1]
\address{Universit\`a degli Studi di Milano,\\ 
Department of Languages, Literatures, Cultures and Mediations,\\
Piazza Sant'Alessandro 1, 20123 Milan, Italy}

\cortext[0]{Please cite as:\\
\textit{M. Ruskov, (2023), ``Grimm in Wonderland: Prompt Engineering with Midjourney to Illustrate
Fairytales'', in: 19th Conference on Information and Research Science Connecting to Digital and
Library Science, CEUR WS vol. 3365, pp. 180–191. URL: \url{https://ceur-ws.org/Vol-3365/paper6.pdf}}.\\}

\begin{abstract}
% Background:
The quality of text-to-image generation is continuously improving, yet the boundaries of its applicability are still unclear. In particular, refinement of the text input with the objective of achieving better results -- commonly called prompt engineering -- so far seems to have not been geared towards work with pre-existing texts.
% Purpose:
We investigate whether text-to-image generation and prompt engineering could be used to generate basic illustrations of popular fairytales.
% Methodology:
Using Midjourney v4, we engage in action research with a dual aim: to attempt to generate 5 believable illustrations for each of 5 popular fairytales, and to define a prompt engineering process that starts from a pre-existing text and arrives at an illustration of it.
% Findings:
We arrive at a tentative 4-stage process: i) initial prompt, ii) composition adjustment, iii) style refinement, and iv) variation selection. We also discuss three reasons why the generation model struggles with certain illustrations: difficulties with counts, bias from stereotypical configurations and inability to depict overly fantastic situations. 
% Practical implications:
Our findings are not limited to the specific generation model and are intended to be generalisable to future ones. 
% Value: 
\end{abstract}

\begin{keywords}
  text-to-image generation \sep
  prompt engineering \sep
  action research \sep
  fairytales  
\end{keywords}

\maketitle

\input{sections/010-intro.tex}
\input{sections/020-litreview.tex}
\input{sections/030-method.tex}
\input{sections/040-analysis.tex}

\input{sections/050-conclusion.tex}

 \begin{acknowledgments}
 The research leading to these results has received funding from the European Union's Horizon 2020 research and innovation programme, in the context of VAST project, under grant agreement No 101004949. This paper reflects only the view of the authors and the European Commission is not responsible for any use that may be made of the information it contains.
 \end{acknowledgments}

%%
%% Define the bibliography file to be used
\bibliography{bibliography}

\begin{center}    
Further example generations available in Appendix on following pages...    
\end{center}

\appendix

\input{sections/060-appendix.tex}

\end{document}

%% file: sections/010-intro.tex
\section{Introduction}
% intro: general setting
Moral values inherent in literary heritage are not explicit and might be perceived differently over time. The project VAST (Values Across Space and Time) sets out to study such variations in perceptions~\cite{castano_computational_2021}. One way to record contemporary perceptions of fairytales is to ask online users what values they are able to identify in text snippets of interest~\cite{ferrara2022ai4ch}. It is challenging to engage people's attention online, and accompanying these questions with illustrations is expected to help improve engagement for participation. 
% challenges
However, pre-existing images are not always readily available for any snippet of interest, and it is impractical to commission ad-hoc illustrations for the purposes of a study where participating users are expected to be exposed to them only for a short period. This opens an opportunity to use a text-to-image generator as a tool to enrich snippets of classical texts for the purposes of improving questionnaire engagement and retention. In turn, this allows for computer-assisted multimedia representation of content that is originally text only, despite apparent limitations, discussed further in this paper.

% CfP: http://lacam.di.uniba.it/IRCDL23/index.php/cfp/

%Applications of Machine Learning Techniques to Research Data and DL
%Cultural Heritage Access - enriching cultural heritage
%Knowledge Representation in Digital Libraries

% goal: what needs to be investigated

In particular, here we set ourselves the task of generating illustrations for fairytales by the Grimm brothers and investigate how accurate we can meet the expectations set by classical texts.
% We focus on some of the most popular fairytales - the ones collected by the Grimm brothers.
% what is already studied  (in contrast to)
While current research into prompt engineering for text-to-image generators focuses typically on construction of creative expressions~\cite{lui2022guidelines, oppenlaender2022ethnographic}, we are rather interested in a believable representation of well-known narratives.
%A bottom-up vs top-down comparison between these approaches could be drawn. In the case of creative investigations, prompts are build from scratch in a bottom-up fashion. Our approach starts top-down from a text snippet and tries to amend it so that the prompt drives the generator to produce at least a partial representation of the original snippet.
% approach: purpose/objective of this study
We engage in an iterative study in the tradition of action research~\cite{staron2020action} while systematically exploring the solution space of text-to-image generation. We set ourselves the exploratory goal to generate at least 5 believable illustrations for each of 5 fairytales and achieve this goal.
% value
This allows us to derive a process-based methodology towards constructing believable representations of preexisting text snippets. We consider our results satisfactory for our purposes of illustrating Grimm's fairytales. Yet, we observe that we have not reached a point where such illustration would be possible for any starting text snippet.

% outline

% Following the typical structure of scientific reports, this introduction is followed by a section on background. Then we present our approach and the resulting method of it's application and conclude with lessons learned and future work.

%% file: sections/020-litreview.tex
\section{Background}

In a work on design guidelines for text-to-image prompts, Liu and Chilton, use the VQGAN+CLIP model and expertiment with 9 prompt templates~\cite{lui2022guidelines}. These templates are phrases in natural language, constructed around up to four building blocks: i) \emph{subject}, ii) \emph{verb}, iii) \emph{medium}, and iv) \emph{style}. In choosing to include medium, they generalise an improvement suggestion by the authors of the generation model. Examples for media that Lui and Chilton provide include \verb|painting|, \verb|photo|, \verb|cartoon|, \verb|icon|, etc.
% Oppenlaender
Oppenlaender used ethnographic methods in their studies~\cite{oppenlaender2022ethnographic}. In the first part of their work, they engaged in an autoethnographic study using VQGAN+CLIP. However this was reported only as a sort of onboarding into the community of prompt engineering practitioners and not a case of reflective practice with its own learned lessons about the process, context, or specific circumstances beyond the conclusions from their second part - the study of other practitioners. In this second, ethnographic part, Oppenalender looks at practices developed in the emerging communities and arrives at a taxonomy of 6 types of prompt modifiers: i) \emph{subject terms}, ii) \emph{style modifiers}, iii) \emph{image prompts}, iv) \emph{quality boosters}, v) \emph{repetition}, and vi) \emph{magic terms}~\cite{oppenlaender2022ethnographic}. Image prompts (i.e. using images as part of the prompt), in particular, are one of the ways practitioners try to enforce character consistency across generations. Notably, the last three prompt modifiers in Oppenlaender's taxonomy are subjectively introduced by practitioners and - due to the randomness of the generation - extremely difficult to validate. 

Whereas the above studies are based on GAN (Generative Adversarial Network), a more recent and well-performing technology is diffusion models~\cite{ulhaq2022diffusion_survey}.
For example, with the DreamBooth model, image-based fine-tuning has been successfully used to enforce character consistency between images~\cite{ruiz2022dreamboot}.
Yet, even in more advanced models, some typical problems persist.
In analysis of one of these -- the Parti model -- its authors list a number of identified typical recurring limitations. Notably, among these are hallucinations, failures with representing counts of similar objects and visual and linguistic priors -- the emergence of stereotypes unrelated to the prompt context~\cite{yu2022parti}. 
% midjourney
Midjourney is among the most popular models among practitioners, even though it is commercial and little is known about its architecture. The current release of Midjourney -- version 4 -- is declared to introduce handling of more complexity, in particular \say{Vastly more knowledge (of creatures, places, and more)}, \say{Much better at getting small details right}, \say{Handles more complex prompting (with multiple levels of detail)}, \say{Better with multi-object / multi-character scenes}~\cite{holz2022midjourney_v4}.
Our preliminary testing showed partial indications that it does deliver on these claims, allegedly on par with with most recent models like Parti~\cite{yu2022parti} and Structured Diffusion Guidance~\cite{feng2022structured}.
% Transfer
However, the studies of prompt engineering listed above focus on a single model and do not give insights as to whether they are transferable across models.
Partly due to the only recent advent of text-to-image generation models that are able to deliver meaningful outputs for complex inputs, systematic attempts at comparison across models are inconclusive~\cite{roberts2022promptVW, ulhaq2022diffusion_survey}. Ideas of how this could be done can come from a related tasks: face generation with GANs, where quantitative comparisons have been made~\cite{borji2022face}.

Due to inherently complex processes, working with a black-box phenomena is very common in social, organisational and design sciences. As a consequence, a range of participatory methods like action research, reflective practice and design research~\cite{mcintosh2010action, staron2020action} are used. Typical for these is that researchers engage in a project as practitioners. In iterative steps they not only develop a product, but also reflect on developing a theory about the task at hand. An intended consequence of this approach is that the emerging theory is contextualised in the specific settings of the project. More specific to action research, two types of learning outcomes are delivered: one intended to be used by practitioners and one by researchers~\cite{staron2020action}. For reasons of space, here we do not report on our implementation of the action research cycle itself, but focus on the applied resulting text illustration process.

%% file: sections/030-method.tex
\section{Method}
% \subsection{Methodological Considerations}
Text-to-image generators share a perceived range of affordances. Not only they take text as input and produce an image, but also let themselves amend with input modifiers. Yet, it remains an open question whether discovered patterns in prompt engineering for one model might be transferable to another. We know that VQGAN+CLIP and Stable Diffusion have very different architectures, and know little of those of Dall-E and Midjourney. Thus, it would be a stretch to assume that the prompt engineering learned for one model would be informative for others.

Instead, we propose that the process of model exploration is a reusable form of knowledge in line with action research. Considering that generation models are black boxes, the experimentation with prompts is much more a field study "in the wild" than a controlled experiment. Thus, we propose that an iterative action research approach could produce knowledge that is more directly transferable across models than phenomenological research into interacting factors.

% \subsection{Approach/Technique}
Having the very specific task of illustrating fairytales, we start from Oppenlaender's taxonomy~\cite{oppenlaender2022ethnographic}. We consider the modifier types of \emph{quality boosters} and \emph{magic terms} to be of little relevance for our task. We also consider \emph{repetition} modifiers out of the scope of this paper. Thus, we focus on \emph{subject terms} and \emph{style modifiers}. Finally, we also take advantage of a feature of Midjourney that allows the creation of variants of a produced image. This can be seen as a special case of \emph{image prompts}.

\paragraph{Subject} Due to our starting point being a pre-existing text and the claimed progress of Midjourney v4, our \emph{subject terms} do not always fit the simple subjects defined e.g. by Liu and Chilton's permutations~\cite{lui2022guidelines}. Rather we derive our subject prompts from the original texts and simplify and adapt them aiming to improve results. A natural first step in this process is to identify where in the original text an important character or moment is introduced. Then we simplify its textual description by trying, whenever possible, to fit it in a simple sentence. In the process we also substitute pronouns with as specific nouns as possible. Examples can be seen in Table~\ref{tab:samples-success}.

\paragraph{Style} We intend style modifiers as a combination of Liu and Chilton's \emph{medium} and \emph{style}. Although these might also not implicitly be necessary for the purposes of our task, we use them to restrict the text-to-image generator. Due to the hallucinations typical for such systems, we seek the possibility to force the generator not to introduce excessive detail which might sharply hit believability. For this purpose we experiment with \emph{style modifiers} like \verb|simple book illustration| or \verb|minimalistic illustration| to restrict hallucinations and lead the generator towards the expectations for the genre medium.

\paragraph{Image prompts} We consider image prompts in a very particular sense, since we do not use the actual possibility to provide a reference image. Instead, we take advantage of an image variation feature provided by Midjourney. Under the premises that it functions conceptually similarly to what an image prompt would be expected to do, even if it is expected to generate results that are much more similar to the reference image than what would be expected from an image prompt. \\

% \subsection{Restrictions}
% consistency across images
Without using image-based fine-tuning, consistency across images is a challenge. in the case of fairytales, it commonly occurs that -- across different generation calls -- the same character is depicted with different features like hair or skin colour. However, for the purposes of this research, we intend to present to users one image at a time, so we do not tackle this issue.
Fine-tuning along the lines of what is done in DreamBooth or actual image prompts, remains beyond the scope of this study.
% assessment
Short time of exposure of the produced images to intended users allows for small inconsistencies between snippet context and image, as long as these do not strongly undermine believability. For the purposes of this preliminary study, we limit ourselves to self-assessing believability.
For the same reasons, we consider five successful image generations per fairytale to be a satisfactory result.
% hi-res
Again, due to the typical model hallucinations, we do not engage in \emph{upscaling} images (increasing resolution), because this inevitably results in further unwanted artefacts. Instead, whenever in future a higher image resolution is necessary, we intend to resort to conventional (basic) resampling techniques.

%% file: sections/040-analysis.tex
\section{Results}\label{sec:results}

% https://docs.google.com/presentation/d/1nMVJEc9ARII9QMppHNvu3OBTOmFtbp87yPPnUN3obFM/edit

In our exploratory study, starting from snippets of text and incrementally refining, we have made more than 650 requests, generating more than 2600 images. Without claiming an efficient exploration, this allowed us to illustrate 5 fairytales with successful generations for at least 5 different snippets per fairytale. Examples of this outcome can be seen in Table~\ref{tab:samples-success} and samples from the steps, preceding these outcomes in Appendix~\ref{appendix:steps}. The remaining successful generations are provided in Appendix~\ref{appendix:all} and the full generated results are available at the author's Midjourney profile\footnote{Consider period from 10/11/2022 to 19/12/2022 at \url{https://www.midjourney.com/app/users/696643755276763136/}, switching to "Grids" from the interface. Free registration in Midjourney is required to get access. Alternatively, contact author for a copy.}. 
From the experience made, we deduce the following tentative four-staged process:

\paragraph{1. Initial prompt} Start with a prompt closely representing the original text trying to summarise it -- preserving vocabulary at this stage -- into as close as possible to a simple sentence.

\paragraph{2. Composition adjustment} Refine prompt step-wise opportunistically, preferring small changes that would allow a fast feedback loop for finer control over change. Pay particular attention to possible misinterpretation to ambiguous words. We identify the possibility to control these at any of three levels:
\begin{itemize}
  \item Adjusting words, optionally simplifying or replacing them with synonyms, ones that might represent the context better. This might include reducing phrasal verbs to one representing the action, sacrificing narrative richness and fidelity for precision of expression.
  \item Add or remove adjectives for entities (subject and objects) or adverbs for verbs
  \item Add objects to represent the context better and/or force removal of unnecessary artefacts.
\end{itemize}

\paragraph{3. Style refinement} Whenever superfluous hallucination of the generator is perceived, it could be suppressed by enforcing a style (in the case of fairytales we propose \verb|illustration|) with modifiers along the lines of \verb|basic|, \verb|simple|, \verb|minimal|, \verb|flatcolor|.

\paragraph{4. Variation selection} Once the desired composition is reached, work with the generation of variants, whenever the generator allows it, as the case of diffusion models like Midjourney. This might be tried also in cases where the composition is only ``nearly reached''. For example, when certain number of objects of a type are needed, but only an approximate count is reached, this step could be attempted in the hope that hallucinations accidentally adjust the count. \\

In our investigation, this produced successful results, a sample of which is provided in Table~\ref{tab:samples-success}.
Even though we indicatively name our steps to describe their primary objective, the corresponding elements are not exclusively elaborated in that step. Rather, one should have low expectations of subsequent steps if the objective of previous steps was not approached to a satisfactory degree. Practitioners are invited to navigate the process freely according to their preferences. On one hand this means that we invite everyone to interrupt it at any step, should the result be considered satisfactory. On the other, we suggest moving back and forth in the process, or even jump steps whenever practitioners see fit.

\input{sections/tab-samples-success.tex}

However, as the samples in Table~\ref{tab:samples-failure} exemplify, the generation of images for other snippets was extremely challenging to produce and we were not successful in doing it. We hypothesise to have identified three particular reasons: difficulties with counts, inability to get distanced from stereotypical configurations and non-conventional situations. These are in line with limitations reported in the Parti model~\cite{yu2022parti}. We subject these hypotheses to simple accessible tests.

There first reason we identify is the difficulty to cause the model to generate a specific number of similar objects. In certain cases this might not be critical. With repeated attempts it is possible to strike three ravens, or it might not be critical if the illustrated dwarfs are five or six, instead of seven. However, a well-known issue among practitioners is the difficulty to draw e.g. hands, often getting a wrong number of fingers.

The second hypothesis is a presumed difficulty to generate scenes, different from a dominant stereotypical view. In previous literature this is typically associated to priors~\cite{yu2022parti} and although in general this could be perceived as an advantage in our task, there are cases when it is undesirable. An examples can be seen in the first row of Table~\ref{tab:samples-failure}. It appears to be impossible to force the creation of a grave without a pre-existing tree on it. Our current hypothesis is that the model "knows" that the grave of Cinderella's mother has a tree on it, as this tree plays an important role further in the story. This hypothesis is put to question by the fact that even when the references of \say{Cinderella} and \say{mother's} are removed, the model continues to produce a tree. We also evaluate how another popular diffusion model behaves on this input. This particular issue was present, but not as persistent when generating with DALL-E.

Examples for failure in representing non-conventional situations could be the extremely poor results for prompts derived from non-realistic texts (also referred to as impossible scenes~\cite{yu2022parti}), such as examples 2 and 3 in Table~\ref{tab:samples-failure}. We hypothesise that this is due to the nature of training based on pre-existing image datasets where anything similar -- albeit possibly present in the dataset -- would be an ignored outlier.

\input{sections/tab-samples-failure.tex}

%% file: sections/tab-samples-success.tex
% Table:
% - original text
% - subject prompt
% - style modifier
% - process stage
% - resulting image
% - believability notes

\def\tabularxcolumn#1{m{#1}}

\begin{table}%[!ht]
\centering
\resizebox{\columnwidth}{!}{
\begin{tabularx}{\textwidth}{XXp{0.03\textwidth}c}
\hline
    \textbf{Original Text} & \textbf{Complete Prompt} & \textbf{Stage} & \textbf{Image} \\
\hline
a little cap made of red velvet.
 Because it suited her so well, and she wanted to wear
 it all the time, she came to be known as
 Little Red Riding Hood
 &
the little cap made of red velvet suited the little
 girl so well, she came to be known as
 Little Red Riding Hood
 &
1
 &
\raisebox{-\totalheight/2}{\includegraphics[width=0.24\textwidth]{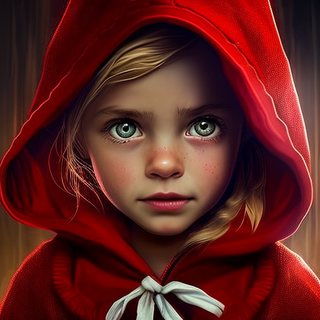}}
 \\
After the full moon had come up... They followed the pebbles that glistened there like newly minted coins, showing them the way
 &
medieval boy and girl follow trace of pebbles in the woods
 &
2
 &
\raisebox{-\totalheight/2}{\includegraphics[width=0.24\textwidth]{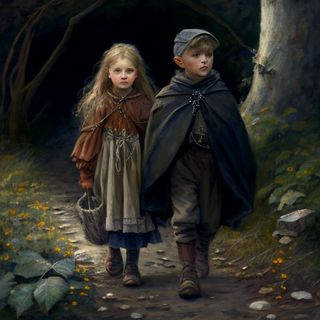}}
 \\
The prince approached her, took her by the hand, and danced with her
 &
The Prince dances with Cinderella, basic book illustration
 &
3
 &
\raisebox{-\totalheight/2}{\includegraphics[width=0.24\textwidth]{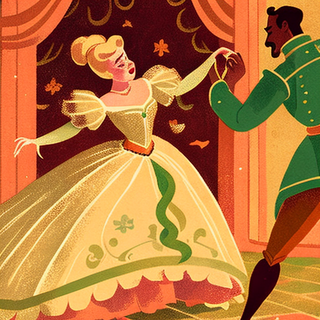}}
 \\
faithful Johannes, who was sitting at the front of the ship making music, saw three ravens flying through the air towards them
 &
three ravens flying by a frigate in open sea, simple book illustration
 &
4
 &
\raisebox{-\totalheight/2}{\includegraphics[width=0.24\textwidth]{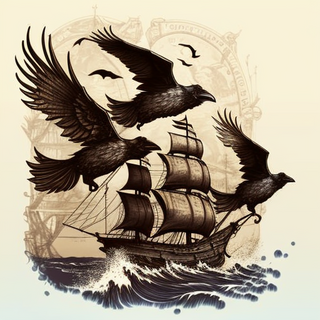}}
 \\
\end{tabularx}
}
\caption{Samples of successful image generations for different fairytales. These represent different stages to show that sometimes satisfactory results can be reached early in the process. Samples from failing steps preceding these successes are shown in Appendix~\ref{appendix:steps}}
\label{tab:samples-success}
\end{table}

%% file: sections/tab-samples-failure.tex
% Table:
% - original text
% - subject prompt
% - style modifier
% - process stage
% - resulting image
% - believability notes

\def\tabularxcolumn#1{m{#1}}

\begin{table}%[t]
\centering
\resizebox{\columnwidth}{!}{
\begin{tabularx}{\textwidth}{XXcc}
\hline
    \textbf{Original Text} & \textbf{Complete Prompt} & \textbf{Midjourney} & \textbf{DALL-E}  \\
 %   \textbf{Text} & \textbf{Prompt} && \\
\hline
Cinderella... went to her mother's grave, and planted the branch on it.
 &
poor girl plants a branch on a grave, minimalistic illustration
 &
\raisebox{-\totalheight/2}{\includegraphics[width=0.24\textwidth]{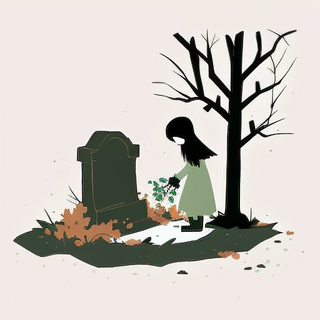}}
 &
\raisebox{-\totalheight/2}{\includegraphics[width=0.24\textwidth]{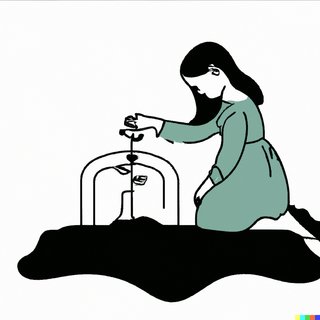}}
 \\
Then Gretel gave [the witch] a shove, causing her to fall in
 &
Gretel shoves the witch into the oven, minimalistic illustration
 &
 \raisebox{-\totalheight/2}{\includegraphics[width=0.24\textwidth]{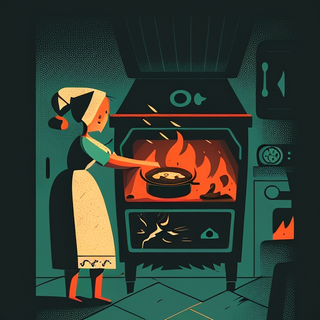}}
 &
 \raisebox{-\totalheight/2}{\includegraphics[width=0.24\textwidth]{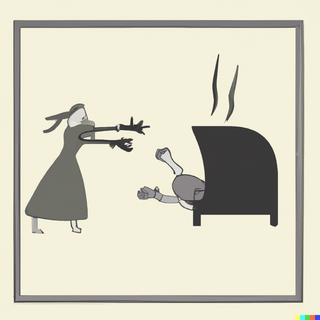}}
 \\
{[the hunter]} took a pair of scissors
and began to cut open the wolf’s belly
 &
the hunter cuts the wolf's belly with scissors, basic illustration
 &
\raisebox{-\totalheight/2}{\includegraphics[width=0.24\textwidth]{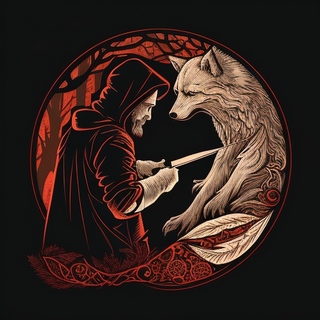}}
 &
\raisebox{-\totalheight/2}{\includegraphics[width=0.24\textwidth]{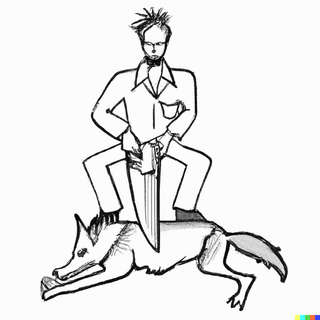}}
 \\
\end{tabularx}
}
\caption{Samples of noticeable failure for image generations, including with alternative generator models. Notice that styles that work well with Midjourney lead to oversimplificated results with DALL-E.}
\label{tab:samples-failure}
\end{table}

%% file: sections/050-conclusion.tex
\section{Future Work}

While our tentative four-staged process was developed and tested with Midjourney v4, we have kept it generic enough to be applicable also to other current generation models and -- most importantly -- future ones to come. This last point is key, because current state-of-the-art models are just arriving at being able to handle a level of complexity required to illustrate an existing text~\cite{feng2022structured, yu2022parti}.
We also have indicated three hypothesised issues for text-to-image generation models, each of which could serve as a challenge for researchers and developers. We claim that an approach starting from intentions, related to a pre-exisitng text, helps shed light onto possible interpretations relevant to model limitations. 

% Future work:
In a subsequent iteration of this action research effort, the domain of studied texts could be further expanded and the exploratory success threshold (the goal) could be increased. While it was not clear whether this would be possible at the start of the study, now we have sufficient confidence to believe that it might be achievable.

As stated by the rationale of this paper, a next step of this research is to perform an usability study with end users to investigate whether generated images actually improve user engagement when responding to online questions about values in fairytales. This study should also include questions about image believability.
Whereas we have tried to limit any bias that the generator might introduce into images, the absence of such bias also needs to be validated. This can be done by comparing responses of end users that are exposed to the generated illustrations with ones that are not.
Finally, we would like to identify metrics that would allow us to measure if user participation corresponds to image quality and believability.

%% file: sections/060-appendix.tex
\clearpage
\section{Samples from Each Step}
\label{appendix:steps}

For each of the examples in Table~\ref{tab:samples-success}, a more detailed illustration of the process is included in Table~\ref{tab:appendix-steps-one}~and~\ref{tab:appendix-steps-two}, featuring examples of previous failing steps in the form of attempted text prompt, resulting images and relevant comments. The images are in a 2x2 grid as produced by Midjourney for any prompt.

\input{sections/tab-appendix1-success-p1.tex}
\input{sections/tab-appendix1-success-p2.tex}

% \section{Other Examples of Above Failures}

% For each of the examples in Table~\ref{tab:samples-failure}, alternative results are provided below to better illustrate the behaviour of the generator in negative cases. Also, we provide a comment explaining why the result presented in Table~\ref{tab:samples-failure} is considered better.

% \input{sections/tab-appendix2-failure.tex}

\clearpage
\section{Remaining Successful Generations}
\label{appendix:all}

For completeness, in the following Tables~\ref{tab:success-all-one}~and~\ref{tab:success-all-two} the remaining successful generations are reported. 

\input{sections/tab-appendix3-all-p1.tex}
\input{sections/tab-appendix3-all-p2.tex}

%% file: sections/tab-appendix1-success-p1.tex
% table
% - original text in caption
% - prompt
% - stage 
% - explanation
% - result
% - comment

\begin{table}[!htbp]%[!ht]
\centering
\resizebox{\columnwidth}{!}{
\begin{tabularx}{\textwidth}{Xp{0.03\textwidth}cX}
\hline
    \textbf{Prompt} & \textbf{Stage} & \textbf{Result} & \textbf{Comments} \\
\hline
% a little cap made of red velvet.
%  Because it suited her so well, and she wanted to wear
%  it all the time, she came to be known as
%  Little Red Riding Hood
%  &
% the little cap made of red velvet suited the little
%  girl so well, she came to be known as
%  Little Red Riding Hood
%  &
% 1
%  &
% \raisebox{-\totalheight/2}{\includegraphics[width=0.24\textwidth]{images/fig3-redcap-s1.png}}
%  \\
Hansel and Gretel follow the pebbles that glisten showing them the way
 &
1
 &
\raisebox{-\totalheight/2}{\includegraphics[width=0.29\textwidth]{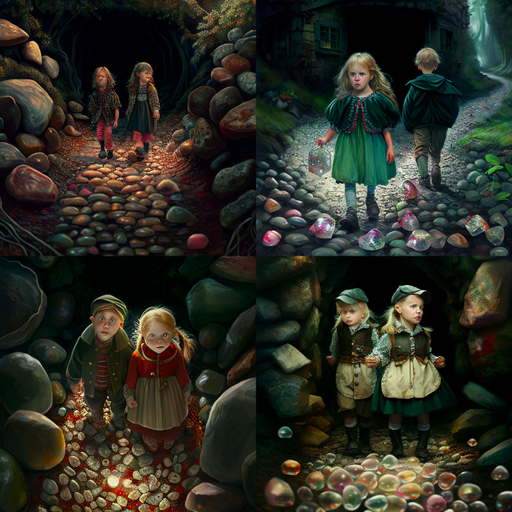}}
 &
\textbf{Problems:} the glistening pebbles do not indicate path; children not representative for Hansel and Gretel; artefacts in faces
 \\ \noalign{\vspace{4pt}}
% medieval boy and girl follow trace of pebbles in the woods
%  &
% 2
%  &
% \raisebox{-\totalheight/2}{\includegraphics[width=0.24\textwidth]{images/fig3-hanselgretel-s2.png}}
%  &
% the occasional white stones could be interpreted as the pebbles in question
%  \\
The Prince dances with Cinderella
&
1
&
\raisebox{-\totalheight/2}{\includegraphics[width=0.29\textwidth]{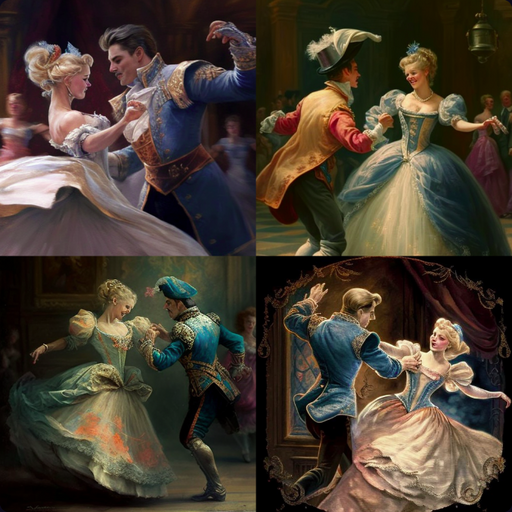}}
&
\textbf{Note:} attempted after reaching Stage 2 where the need to simplify the prompt was identified. 

\textbf{Problems:} artefacts in fingers and faces
 \\ \noalign{\vspace{4pt}}
The Prince came to Cinderella, and took her by the hand and danced with her &
2
 &
\raisebox{-\totalheight/2}{\includegraphics[width=0.29\textwidth]{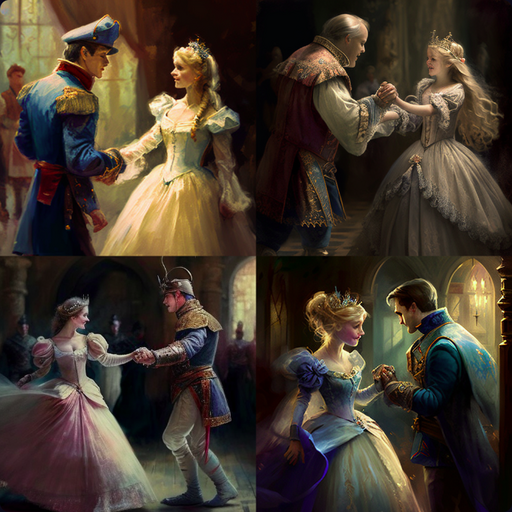}}
&
\textbf{Problems:} artefacts in hands and faces
\end{tabularx}
}
\caption{Samples from stages that led up to the results for Hansel and Gretel and Cinderella in Table~\ref{tab:samples-success}.}
\label{tab:appendix-steps-one}
\end{table}

%% file: sections/tab-appendix1-success-p2.tex
% table
% - original text in caption
% - prompt
% - stage 
% - explanation
% - result
% - comment

\begin{table}[!htbp]%[!ht]
\centering
\resizebox{\columnwidth}{!}{
\begin{tabularx}{\textwidth}{Xp{0.03\textwidth}cX}
\hline
    \textbf{Prompt} & \textbf{Stage} & \textbf{Result} & \textbf{Comments} \\
\hline
Johannes on ship playing music and three ravens flying &
1
 &
\raisebox{-\totalheight/2}{\includegraphics[width=0.29\textwidth]{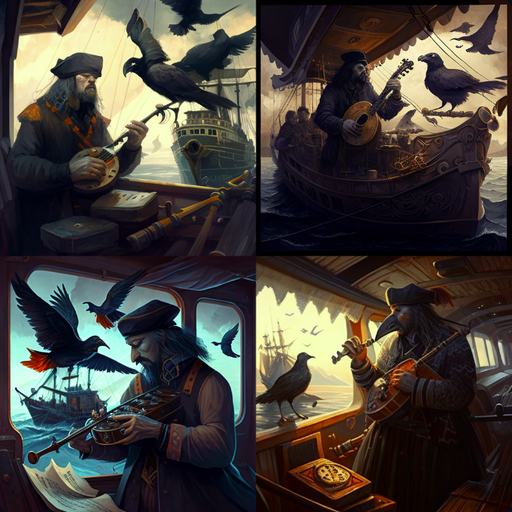}}
&
\textbf{Problems:} the context of Johannes hiding is missing; number of ravens; ships in background might be misleading
 \\ \noalign{\vspace{4pt}}
three ravens flying and faithful Johannes hiding on ship
&
2
 &
\raisebox{-\totalheight/2}{\includegraphics[width=0.29\textwidth]{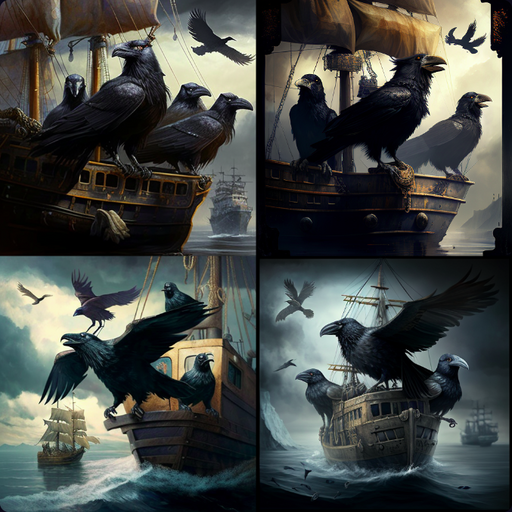}}
&
\textbf{Note:} hiding Johannes is actually not visible.

\textbf{Problems:} Number of ravens; strange ships
 \\ \noalign{\vspace{4pt}}
three ravens flying by the royal frigate in open sea, simple book illustration
&
3
 &
\raisebox{-\totalheight/2}{\includegraphics[width=0.29\textwidth]{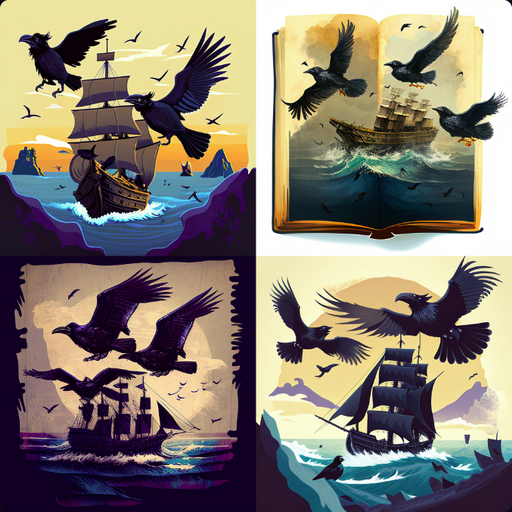}}
&
\textbf{Problems:} number of ravens; ``book'' from style showing in image
\end{tabularx}
}
\caption{Samples from all stages that led up to the result for Faithful Johannes in Table~\ref{tab:samples-success}.}
\label{tab:appendix-steps-two}
\end{table}

%% file: sections/tab-appendix3-all-p1.tex
\begin{table}[!hbp]%[!ht]
\centering
% \resizebox{\columnwidth}{!}{

\input{sections/tab-appendix3-redcap.tex}
\input{sections/tab-appendix3-cinderella.tex}

% }
\caption{The generations that are considered to be successful beyond the ones shown in Table~\ref{tab:samples-success}. Here for Little Red Riding Hood and Cinderella. A white border is used to indicate which image of the 2x2 grid was considered successful.}
\label{tab:success-all-one}
\end{table}

%% file: sections/tab-appendix3-redcap.tex
 % \def\tabularxcolumn#1{m{#1}}

% \begin{table}[!htbp]%[!ht]
% \centering
% \resizebox{\columnwidth}{!}{
\begin{tabularx}{\textwidth}{XXXX}%{XXp{0.03\textwidth}c}
% \hline
%     \textbf{Original Text} & \textbf{Complete Prompt} & \textbf{Stage} & \textbf{Image} \\
% \hline
\raisebox{-\totalheight/2}{\includegraphics[width=0.24\textwidth]{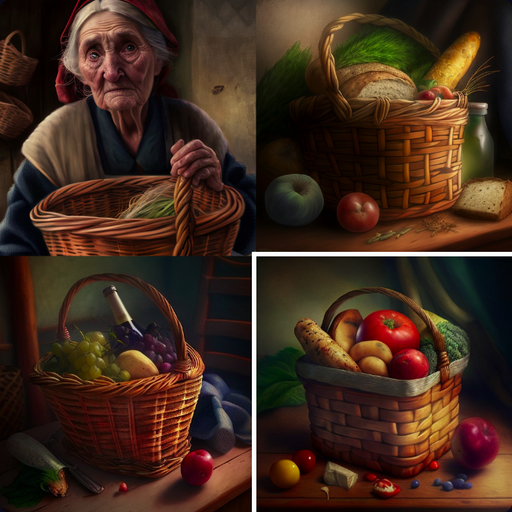}}
 &
\raisebox{-\totalheight/2}{\includegraphics[width=0.24\textwidth]{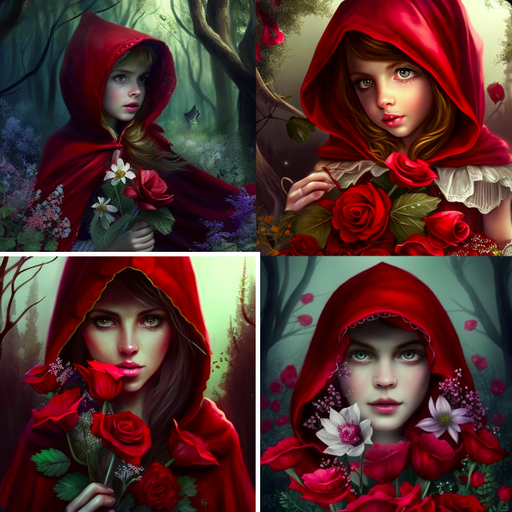}}
 &
\raisebox{-\totalheight/2}{\includegraphics[width=0.24\textwidth]{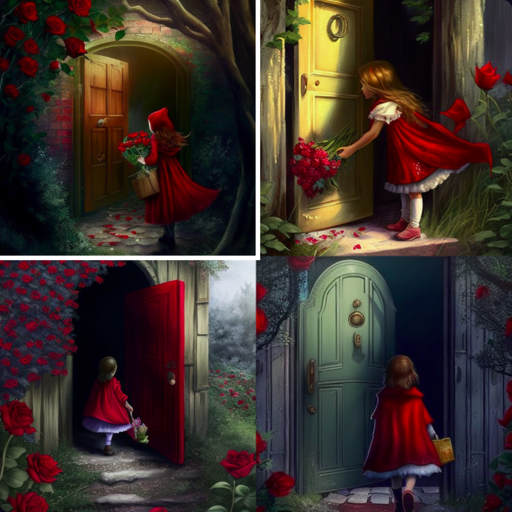}}
 &
\raisebox{-\totalheight/2}{\includegraphics[width=0.24\textwidth]{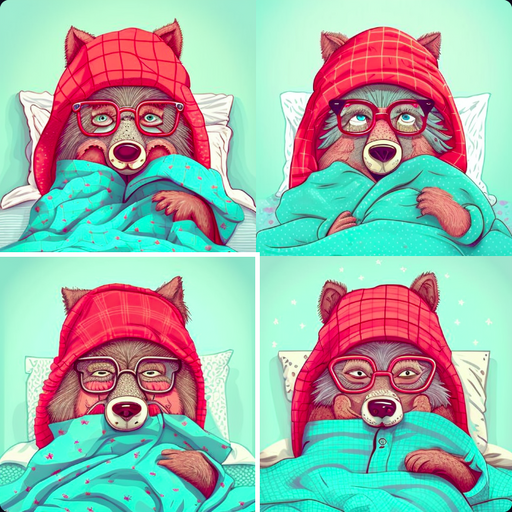}}
 \\
\textbf{Original:} Take them to your grandmother. She is sick and weak, and they will do her well.
 
\textbf{Prompt:} your grandmother is sick and weak, and the food in this basket will do her well
 &
\textbf{Original and prompt:} Little Red Riding Hood, haven't you seen the beautiful flowers that are blossoming in the woods
 &
\textbf{Original and prompt:} Little Red Riding Hood had run after flowers, and did not continue on her way to grandmother's until she had gathered all that she could carry. When she arrived, she found, to her surprise, that the door was open
 &
\textbf{Original:} Then he took [granny's] clothes, put them on, and put her cap on his head. He got into her bed and pulled the curtains shut.
 
\textbf{Prompt:} The big bad wolf put on granny's clothes, her cap on his head and her glasses. Then he got into her bed and under her duvet, flat color en face
\end{tabularx}
% }
% \caption{Successful generations for Little Red Riding Hood.}
% \label{tab:redcap-all}
% \end{table}

%% file: sections/tab-appendix3-cinderella.tex
%  \def\tabularxcolumn#1{m{#1}}

% \begin{table}[!htbp]%[!ht]
% \centering
% \resizebox{\columnwidth}{!}{
\begin{tabularx}{\textwidth}{XXXX}%{XXp{0.03\textwidth}c}
% \hline
%     \textbf{Original Text} & \textbf{Complete Prompt} & \textbf{Stage} & \textbf{Image} \\
% \hline
\raisebox{-\totalheight/2}{\includegraphics[width=0.24\textwidth]{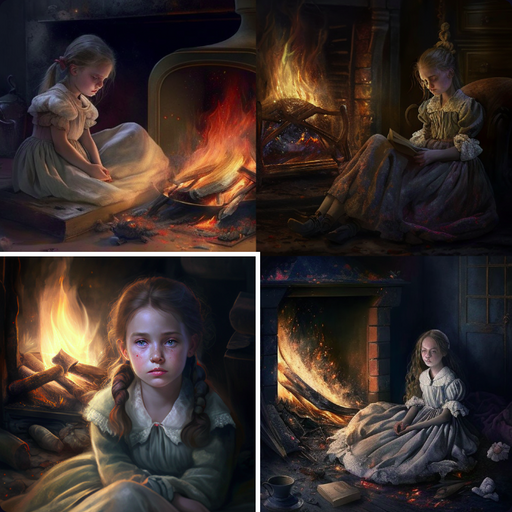}}
 &
\raisebox{-\totalheight/2}{\includegraphics[width=0.24\textwidth]{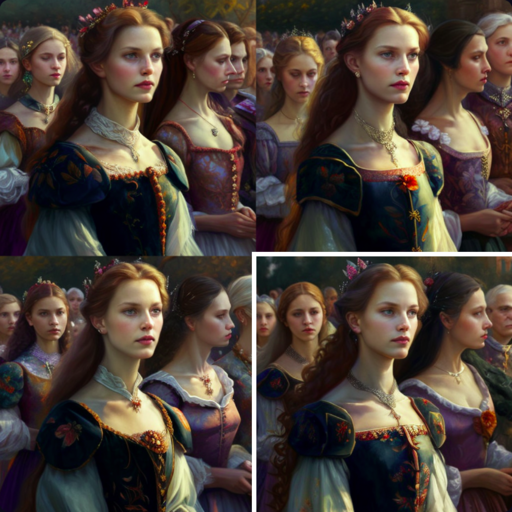}}
 &
\raisebox{-\totalheight/2}{\includegraphics[width=0.24\textwidth]{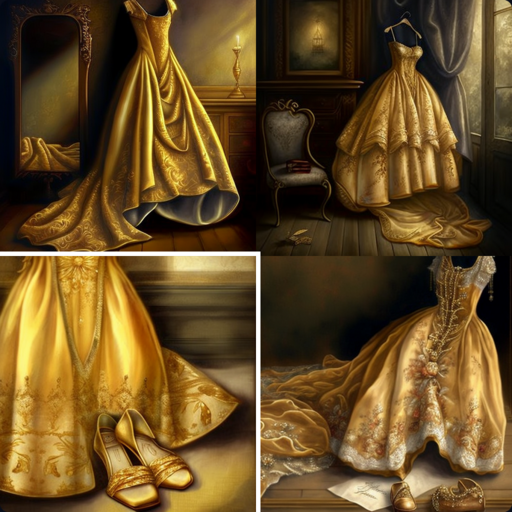}}
 &
\raisebox{-\totalheight/2}{\includegraphics[width=0.24\textwidth]{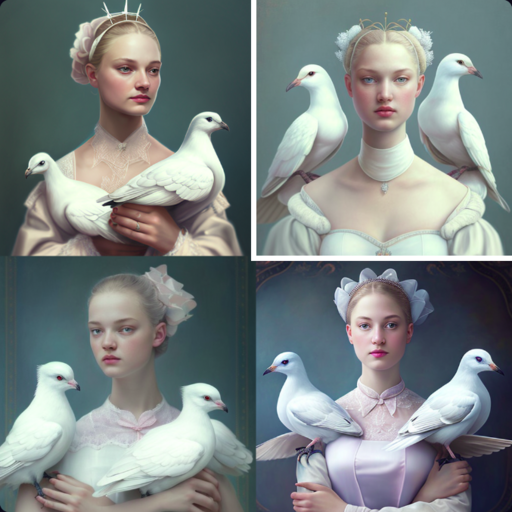}}
 \\
\textbf{Original:} there was no bed for her... she had to sleep by the fireside in the ashes
 
\textbf{Prompt:} Cinderela had no bed and had to sleep by the fireside in the ashes
 &
\textbf{Original:} the king proclaimed a festival... All the beautiful young girls... were invited, so that the prince could select a bride for himself
 
\textbf{Prompt:} a festival to which all beautiful young girls are invited, so that the prince might choose
 &
\textbf{Original and prompt:} dress that was more splendid and magnificent than any she had yet had, and the slippers were of pure gold
 &
\textbf{Original:} After [the pigeons] had cried this out, they both flew down and perched on Cinderella's shoulders, one on the right, the other on the left, and remained sitting there.
 
\textbf{Prompt:} the white doves perched on each of Cinderella's shoulders, simple
\end{tabularx}
% }
% \caption{Successful generations for Cinderella.}
% \label{tab:redcap-all}
% \end{table}

%% file: sections/tab-appendix3-all-p2.tex
\begin{table}[!htbp]%[!ht]
\centering
% \resizebox{\columnwidth}{!}{

\input{sections/tab-appendix3-hanselngretel.tex}
\input{sections/tab-appendix3-trustyjohn.tex}
\input{sections/tab-appendix3-snowwhite.tex}

% }
\caption{The generations that are considered to be successful beyond the ones shown in Table~\ref{tab:samples-success}. Here for Hansel and Gretel, Faithful Johannes and Little Snow White. A white border indicates the successful generation.}
\label{tab:success-all-two}
\end{table}

%% file: sections/tab-appendix3-hanselngretel.tex
 % \def\tabularxcolumn#1{m{#1}}

% \begin{table}[!htbp]%[!ht]
% \centering
% \resizebox{\columnwidth}{!}{
\begin{tabularx}{\textwidth}{XXXX}%{XXp{0.03\textwidth}c}
% \hline
%     \textbf{Original Text} & \textbf{Complete Prompt} & \textbf{Stage} & \textbf{Image} \\
% \hline
\raisebox{-\totalheight/2}{\includegraphics[width=0.24\textwidth]{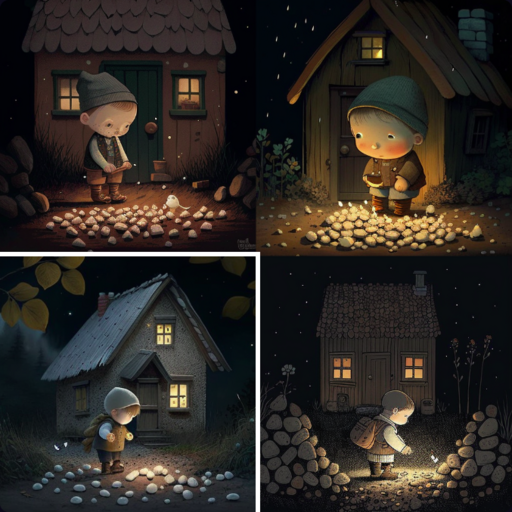}}
 &
\raisebox{-\totalheight/2}{\includegraphics[width=0.24\textwidth]{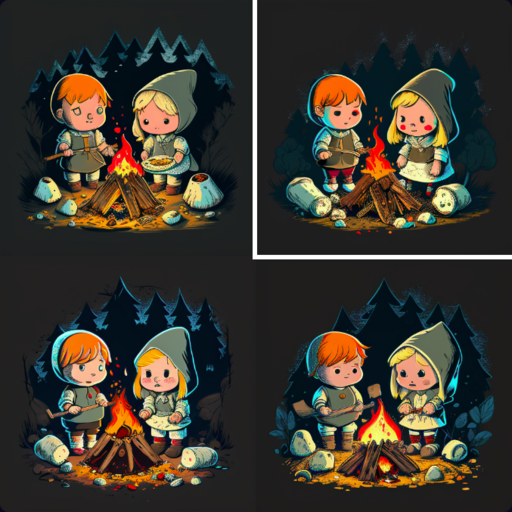}}
 &
\raisebox{-\totalheight/2}{\includegraphics[width=0.24\textwidth]{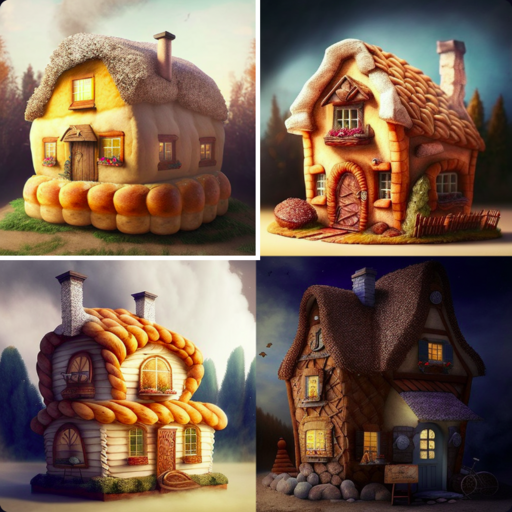}}
 &
\raisebox{-\totalheight/2}{\includegraphics[width=0.24\textwidth]{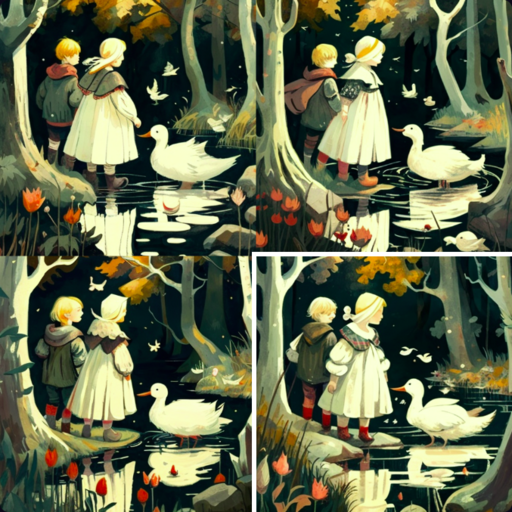}}
 \\
\textbf{Original:} The moon was shining brightly... Hansel bent over and filled his jacket pockets with [pebbles], as many as would fit.
 
 \textbf{Prompt:} little Hansel fills his pockets with white pebbles in front of woodcutter's house at night, simple illustration
 &
 \textbf{Original:} Hansel and Gretel sat by the fire... each one ate his little piece of bread.
 
 \textbf{Prompt:} poor Hansel and Gretel hold only breadcrumbs by campfire, simple illustration
 &
 \textbf{Original and prompt:} the little house was built entirely from bread with a roof made of cake, and the windows were made of clear sugar
 &
 \textbf{Original:} they arrived at a large body of water...  [Gretel says] ``there is a white duck swimming''
 
 \textbf{Prompt:} boy and girl see white duck swimming in a lake In the forest, medieval illustration
\end{tabularx}
% }
% \caption{Successful generations for Hansel and Gretel.}
% \label{tab:redcap-all}
% \end{table}

%% file: sections/tab-appendix3-trustyjohn.tex
 % \def\tabularxcolumn#1{m{#1}}

% \begin{table}[!htbp]%[!ht]
% \centering
\begin{tabularx}{\textwidth}{XXXX}%{XXp{0.03\textwidth}c}
% \hline
%     \textbf{Original Text} & \textbf{Complete Prompt} & \textbf{Stage} & \textbf{Image} \\
% \hline
\raisebox{-\totalheight/2}{\includegraphics[width=0.24\textwidth]{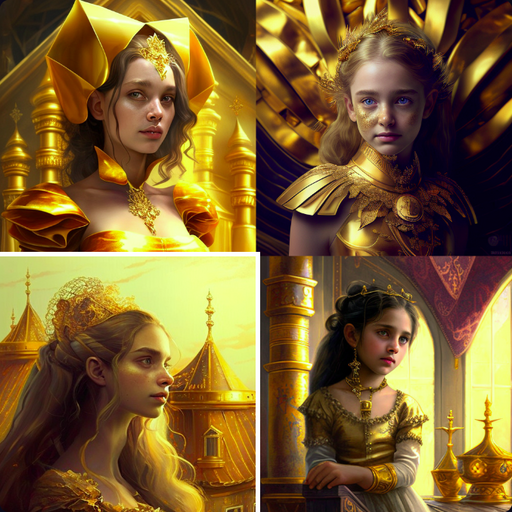}}
 &
\raisebox{-\totalheight/2}{\includegraphics[width=0.24\textwidth]{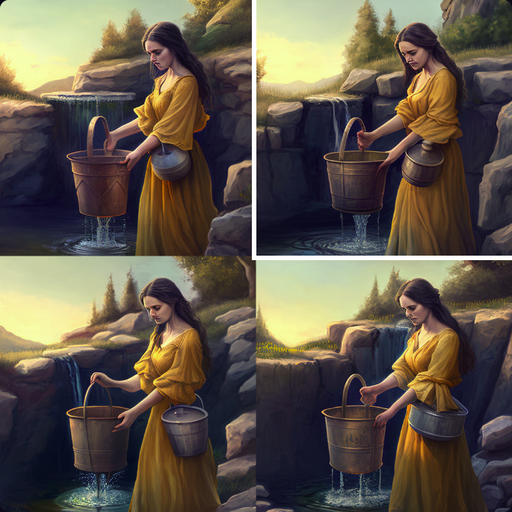}}
 &
\raisebox{-\totalheight/2}{\includegraphics[width=0.24\textwidth]{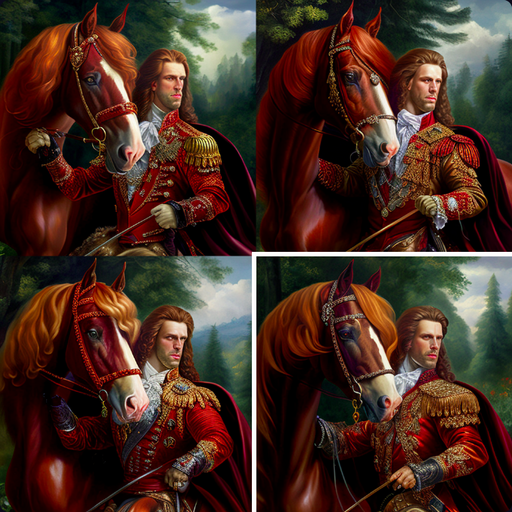}}
 &
\raisebox{-\totalheight/2}{\includegraphics[width=0.24\textwidth]{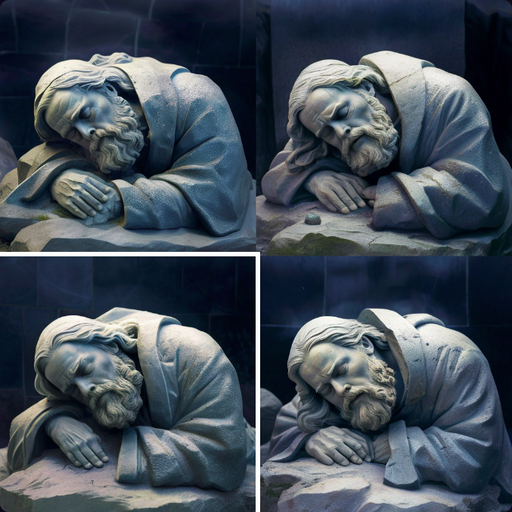}}
 \\
\textbf{Original:} [Johannes says] ``Everything which [the princess] has about her is of gold]''
 
\textbf{Prompt:} Everything the princess of the golden roof has about her is of gold
 &
\textbf{Original and prompt:} a beautiful girl was standing there by the well with two golden buckets in her hand, drawing water with them
 &
\textbf{Original:} a magnificent chestnut horse sprang forward... He was about to mount it...
  
\textbf{Prompt:} the king wants to mount beautiful chestnut horse
 &
\textbf{Original:} But as faithful Johannes spoke the last word, he fell down lifeless and turned to stone
 
\textbf{Prompt:} Faithful Johannes turns into lifeless stone
\end{tabularx}
% }
% \caption{Successful generations for Faithful Johannes.}
% \label{tab:redcap-all}
% \end{table}

%% file: sections/tab-appendix3-snowwhite.tex
%  \def\tabularxcolumn#1{m{#1}}

% \begin{table}[!htbp]%[!ht]
% \centering
% \resizebox{\columnwidth}{!}{
\begin{tabularx}{\textwidth}{XXXX}%{XXp{0.03\textwidth}c}
% \hline
%     \textbf{Original Text} & \textbf{Complete Prompt} & \textbf{Stage} & \textbf{Image} \\
% \hline
\raisebox{-\totalheight/2}{\includegraphics[width=0.24\textwidth]{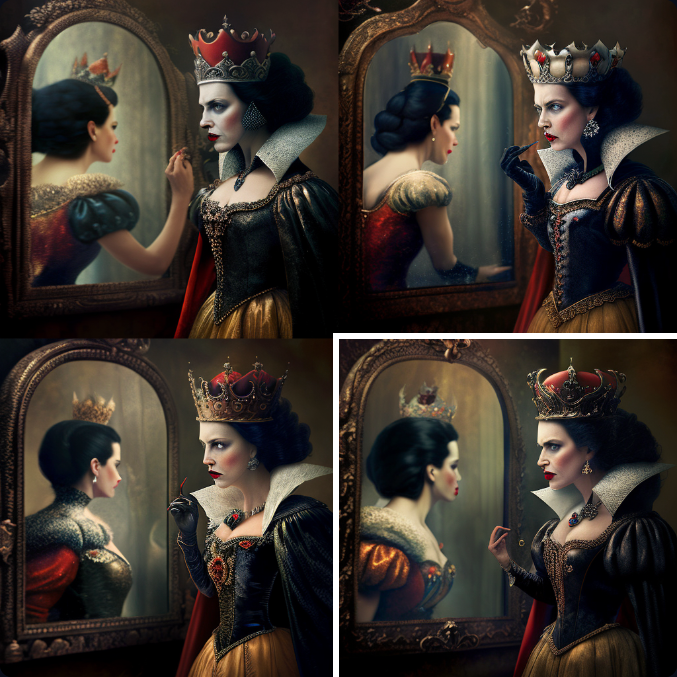}}
 &
\raisebox{-\totalheight/2}{\includegraphics[width=0.24\textwidth]{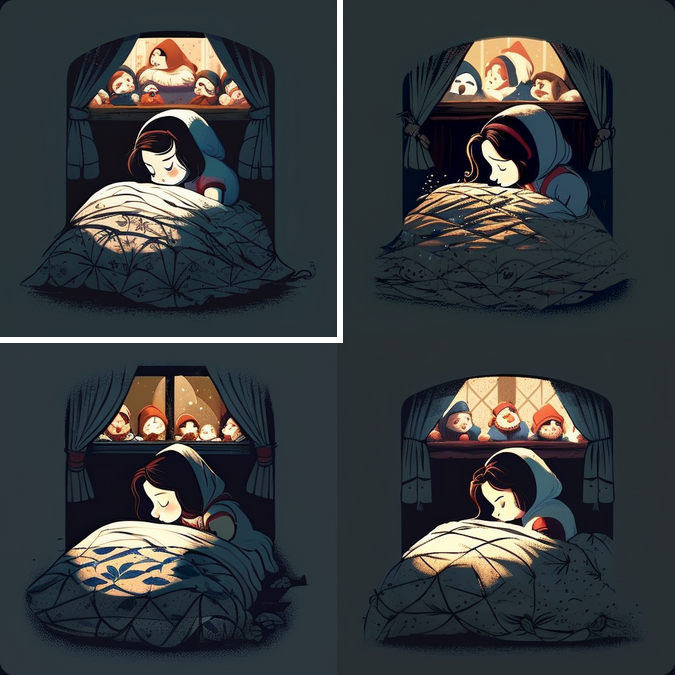}}
 &
\raisebox{-\totalheight/2}{\includegraphics[width=0.24\textwidth]{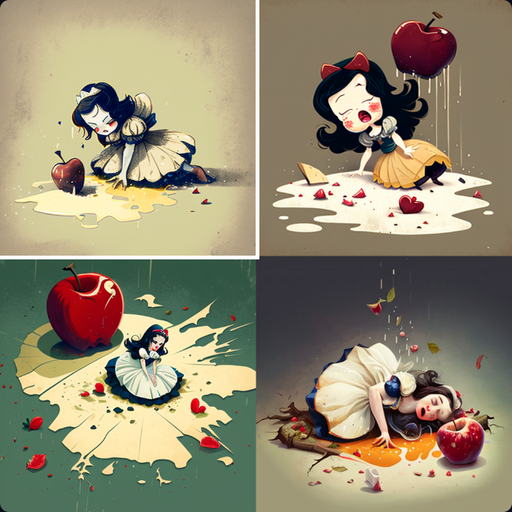}}
 &
\raisebox{-\totalheight/2}{\includegraphics[width=0.24\textwidth]{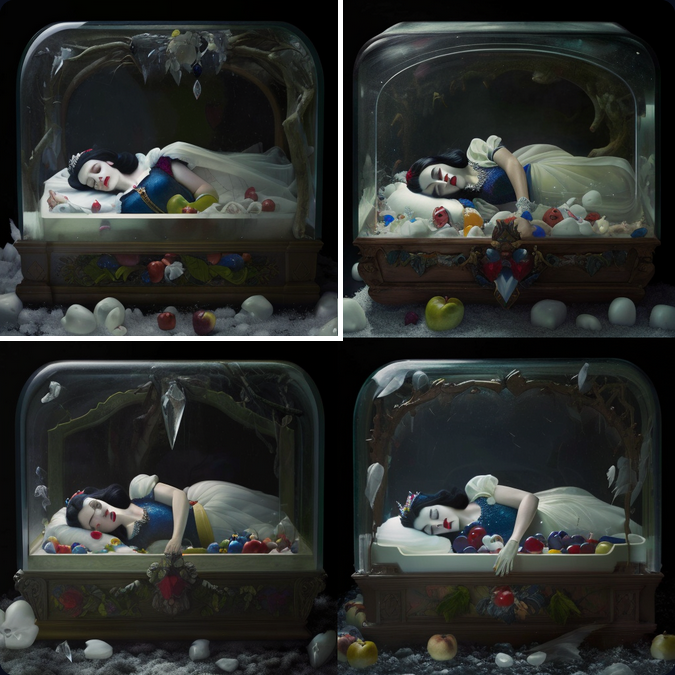}}
 \\
\textbf{Original:} ``Mirror, mirror, on the wall, Who in this land is fairest of all?''
``It answered:... Snow-White is a thousand times fairer than you.''
 
\textbf{Prompt:} evil queen sees Snowwhite in mirror mirror on the wall
 &
\textbf{Original:} [a dwarf] found Snow-White lying [in his bed] asleep. The seven dwarfs all came running up
 
\textbf{Prompt:} Beautiful Snowwhite sleeps in bed and drawves watch her, minimalistic illustration
 &
 \textbf{Original:} [Snow White] barely had a bite [of the apple] in her mouth when she fell to the ground dead.
 
\textbf{Prompt:} Snowwhite collapses on ground and drops apple, basic illustration
 &
 \textbf{Original:} they had a transparent glass coffin made, so she could be seen from all sides
 
\textbf{Prompt:} Snowwhite laying dead in a glass coffin
\end{tabularx}
% }
% \caption{Successful generations for Snow White.}
% \label{tab:redcap-all}
% \end{table}